\documentclass{article}
\usepackage[preprint]{neurips_2026}

\usepackage[utf8]{inputenc}
\usepackage[T1]{fontenc}
\usepackage{hyperref}
\usepackage{url}
\usepackage{booktabs}
\usepackage{amsfonts}
\usepackage{nicefrac}
\usepackage{xcolor}
\usepackage{amsmath,amssymb,amsfonts}
\usepackage{mathtools}
\usepackage{amsthm}
\usepackage{textcomp}
\usepackage{mathrsfs}
\usepackage{algorithm}
\usepackage{algorithmic}
\usepackage{placeins}
\usepackage{comment}
\usepackage{tikz}
\usepackage{tabularx}
\usepackage{graphicx}
\usepackage{subcaption}
\usepackage{wrapfig}
\usepackage[capitalize,noabbrev]{cleveref}

\theoremstyle{plain}
\newtheorem{theorem}{Theorem}[section]
\newtheorem{proposition}[theorem]{Proposition}

\theoremstyle{definition}

\theoremstyle{remark}

\title{Synthetic Counterfactual Labels for Efficient Conformal Counterfactual Inference}

\author{%
  Amirmohammad Farzaneh \\
  Institute for Intelligent Networked Systems (INSI) \\
  Northeastern University London \\
  London, UK \\
  \texttt{a.farzaneh@northeastern.edu} \\
  \And
  Matteo Zecchin \\
  Communication Systems Department\\
  EURECOM \\
  Sophia Antipolis, France \\
  \texttt{zecchin@eurocom.fr} \\
  \And
  Osvaldo Simeone \\
  Institute for Intelligent Networked Systems (INSI) \\
  Northeastern University London \\
  London, UK \\
  \texttt{o.simeone@northeastern.edu} \\
}

\begin{document}

\maketitle

\begin{abstract}
This work addresses the problem of constructing reliable prediction intervals for individual counterfactual outcomes. Existing conformal counterfactual inference (CCI) methods provide marginal coverage guarantees but often produce overly conservative intervals, particularly under treatment imbalance when counterfactual samples are scarce. We introduce synthetic data-powered CCI (SP-CCI), a new framework that augments the calibration set with synthetic counterfactual labels generated by a pre-trained counterfactual model. To ensure validity, SP-CCI incorporates synthetic samples into a conformal calibration procedure based on  risk-controlling prediction sets (RCPS) with a debiasing step informed  by prediction-powered inference (PPI). We prove that SP-CCI achieves tighter prediction intervals while preserving marginal coverage, with theoretical guarantees under both exact and approximate importance weighting. Empirical results on different datasets confirm that SP-CCI consistently reduces interval width compared to standard CCI across all settings.
\end{abstract}

\section{Introduction}
\label{sec:intro}

\subsection{Context and Motivation}
\label{sec:context}

Consider a medical decision-making scenario in which a clinician must decide whether to administer a costly treatment, such as a new cancer therapy, to a patient. Each patient is characterized by a set of covariates $X$, e.g., demographics, medical history, and diagnostic test results, and may receive either the treatment ($T=1$) or no treatment ($T=0$). The observed outcome $Y^{\text{obs}} = Y(T)$ could be a clinical metric such as tumor size reduction. The \emph{counterfactual} outcome $Y^{\text{cf}} = Y(1-T)$ represents what would have happened had the patient received the other treatment option.

\emph{Individual} counterfactual outcomes are fundamental to treatment effect estimation and policy evaluation. Clinicians are often interested not only in whether a new therapy outperforms standard care on average, but also in how much benefit it offers for a specific patient or for subgroups defined by covariates $X$.  Achieving this goal requires quantifying uncertainty in predictions of the unobserved counterfactual outcome $Y^{\text{cf}}$. The challenge lies in the fundamental \emph{missing data problem}: for each patient, only one of the potential outcomes $(Y(0), Y(1))$, namely $Y^{\text{obs}}$, is observed,
 while the corresponding counterfactual outcome $Y^{\text{cf}}$ is never directly available.

A promising solution to the problem of uncertainty quantification for individual counterfactual outcomes comes from \emph{conformal prediction} \citep{shafer2008tutorial, vovk2005algorithmic}, a post-hoc calibration method that provides statistically valid prediction intervals without strong distributional assumptions. The recent technique  introduced in \citep{lei2021conformal}, referred to here as \emph{conformal counterfactual inference} (CCI), adapts conformal prediction to construct prediction intervals for counterfactual outcomes with guaranteed marginal coverage. These guarantees hold regardless of the accuracy of the underlying predictive model.

\begin{figure}[h]
    \centering
    \includegraphics[width=0.70\linewidth]{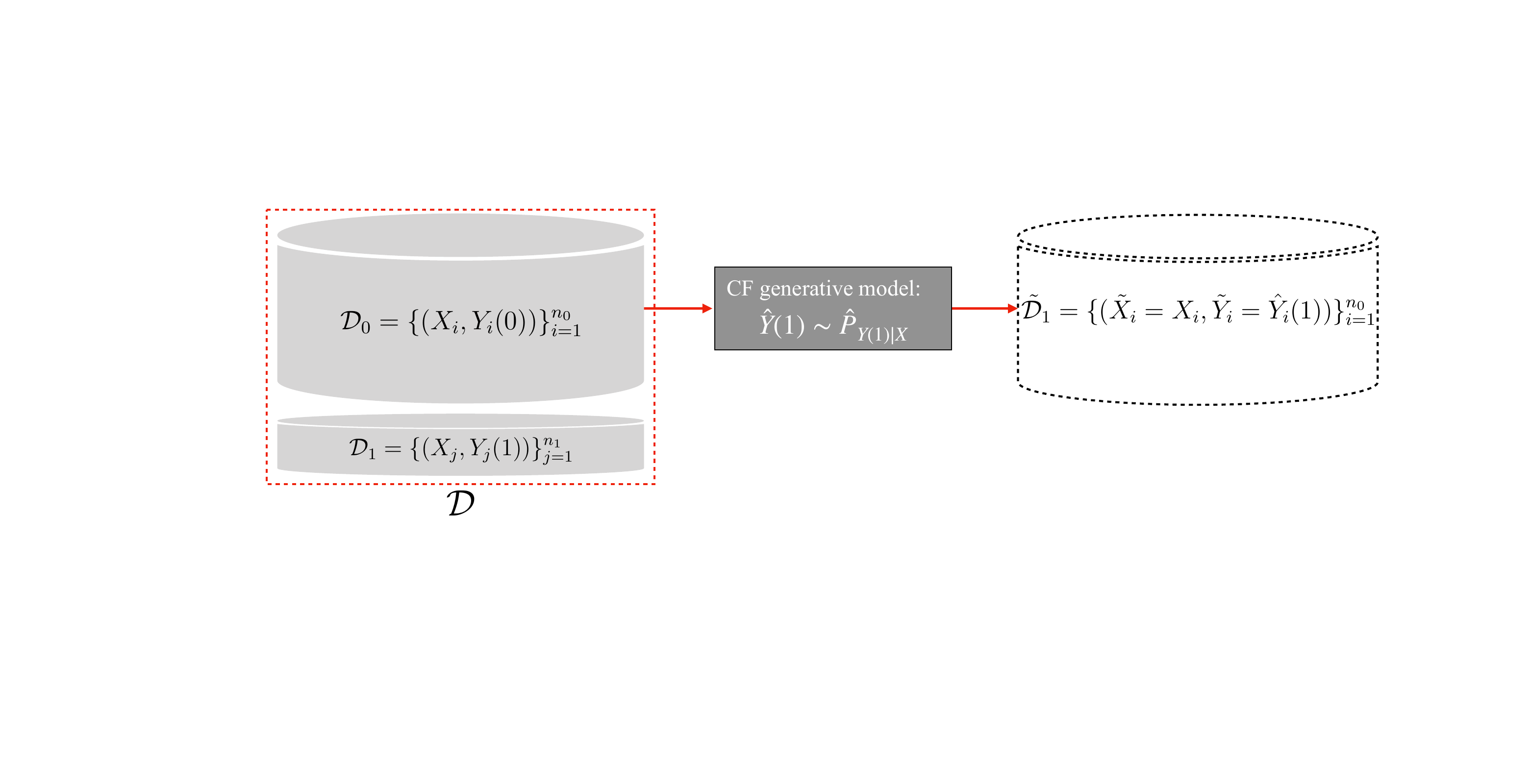}
    \caption{The proposed synthetic data-powered conformal counterfactual inference (SP-CCI) method leverages synthetic counterfactual labels $\hat{Y}(1)$ produced using a pre-trained generative model $\hat{P}_{Y(1)|X}$ from the, typically larger, dataset $\mathcal{D}_0$ ($n_0 \gg n_1$).}
    \label{fig:data}
\end{figure}

CCI requires \textit{calibration data} encompassing observations from the treatment arm whose outcome we wish to predict. However, in many medical datasets, treatment assignment is highly imbalanced: expensive or experimental treatments are administered only to a small fraction of patients, resulting in very few calibration samples for that arm \citep{dahabreh2020extending}. As an illustration, in Fig.~\ref{fig:data}, the data set reporting the outcome $Y(0)$, corresponding to the control group, is larger than that reporting the treatment outcomes $Y(1)$.

Similar imbalances occur beyond medicine: in online advertising with sparse exposure to ad variants \citep{bottou2013counterfactual}, in recommendation systems where many items lack interaction data \citep{swaminathan2015counterfactual}, and in A/B testing where risky variants are shown to only a few users \citep{kohavi2020trustworthy}. In all cases, scarce calibration data for the target arm yields wide, uninformative intervals, limiting CCI's utility in high-stakes decisions.

A natural idea is to use \emph{synthetic} counterfactual outcomes generated by an existing pre-trained model. Such models can produce plausible counterfactual outcomes for patients in the larger control group, increasing the effective calibration sample size for the treatment arm. However, directly including the resulting synthetic data points in the CCI procedure breaks its statistical validity in terms of marginal coverage, since synthetic outcomes are biased approximations of the truth.

In this context, we propose \emph{synthetic data-powered CCI} (SP-CCI), a conformal counterfactual inference framework that addresses the data imbalance problem by augmenting the calibration set with synthetic counterfactual labels, while preserving marginal coverage guarantees. SP-CCI integrates synthetic and real calibration data through a debiased miscoverage estimator informed by (\emph{i}) prediction-powered inference (PPI) \citep{angelopoulos2023prediction}, which corrects for bias introduced by approximate labels and (\emph{ii}) risk-controlling prediction sets (RCPS) \citep{bates2021distribution}, which choose the smallest interval widening that ensures the miscoverage risk is controlled with high probability. The result is a method that preserves CCI's validity while improving its efficiency, yielding narrower prediction intervals, especially when high-quality synthetic counterfactual generators are available.

\textbf{Scope:} Following \citep{lei2021conformal}, SP-CCI targets prediction intervals for a single counterfactual potential outcome. Specifically, for a test unit with factual treatment $T=0$ and observed outcome $Y(0)$, the goal is to construct a valid interval for the unobserved counterfactual $Y(1)$. Unlike \citep{lei2021conformal}, SP-CCI leverages calibration data from the untreated arm, requiring a modified form of importance 
weighting. Note that, constructing valid intervals when \emph{neither} potential outcome is observed is a generally harder open problem studied separately in \citep{lei2021conformal,alaa2023conformal}. We leave extensions of SP-CCI to this setting for future work.

\textbf{Coverage guarantee:} SP-CCI provides a \emph{high-probability} marginal coverage guarantee, controlled by a user-specified probability parameter $\delta$. This differs from CCI's \emph{unconditional} marginal coverage guarantee; we discuss this distinction and its practical implications in Sec.~\ref{sec:theoretical_guarantees}.

\subsection{Further Related Work and Main Contributions}
\label{sec:related}

A detailed discussion of related work on counterfactual inference, conformal prediction for counterfactuals, and synthetic data in causal inference is provided in Appendix~\ref{appendix:related_work}. We highlight here the GESPI framework \citep{bashari2026statistical}, which provides distribution-free inference leveraging synthetic data. Accordingly, it could be used as an alternative to PPI to construct prediction intervals for the counterfactual setting studied here.

The main contributions of this work are summarized as follows.

    $\bullet$ \textbf{Methodology:} We introduce SP-CCI, a conformal counterfactual inference method that combines real and synthetic calibration data via a debiased miscoverage estimator, ensuring valid high-probability coverage.

    $\bullet$ \textbf{Theory:} We provide formal coverage guarantees under exact and approximate importance weighting, quantifying the effect of weight misspecification.

    $\bullet$ \textbf{Applications:} We evaluate SP-CCI on synthetic data \citep{lei2021conformal} and on the semi-synthetic IHDP dataset \citep{hill2011bayesian}, showing consistent efficiency gains over CCI \citep{lei2021conformal} and GESPI \citep{bashari2026statistical}.

\section{Problem Definition}
\label{sec:problem_definition}

We consider the standard potential outcome framework for counterfactual inference with a binary treatment \citep{splawa1990application, rubin1974estimating}. Specifically, each unit $i$ is associated with an observed covariate $X_i \in \mathcal{X}$, the pair of potential outcomes $(Y_i(1), Y_i(0))$, and a binary treatment assignment $T_i \in \{0,1\}$ determining the observed outcome $Y_i^{\text{obs}}$. Under the \textit{stable unit treatment value assumption} (SUTVA) \citep{rubin1990formal}, the observed outcome $Y_i^{\text{obs}}$ is the potential outcome under the treatment $T_i$ and is given by $Y_i^{\text{obs}} = Y_i(T_i)$. The unobserved outcome $Y_i^{\text{cf}}$, also referred to as the \textit{counterfactual} outcome, is the potential outcome under treatment not received, i.e., $Y_i^{\text{cf}} = Y_i(1-T_i)$. Thus, for each unit $i$, we observe the triplet \( (X_i, T_i, Y_i^{\text{obs}}) \), which includes only one of the potential outcome $Y_i(T_i)$ corresponding to the treatment $T_i$.

We assume that the pair of potential outcomes, treatment, and covariate for each unit $i$ are drawn independently and identically from a joint distribution $P_{Y(0), Y(1), T, X}$, i.e.,
\begin{equation}
\label{eq:joint}
(Y_i(1), Y_i(0), T_i, X_i) \stackrel{\text{i.i.d.}}{\sim} P_{Y(1), Y(0), T, X}.
\end{equation}
Throughout, we adopt the standard assumption of \textit{strong ignorability} \citep{rosenbaum1983central, rubin1978bayesian, imbens2015causal}. This asserts that, under the given joint distribution in (\ref{eq:joint}), the assigned treatment is conditionally independent of the potential outcomes given the covariates, i.e.,
\begin{equation}
\label{eq:ignorability}
P_{Y(1), Y(0), T|X}=P_{Y(1), Y(0)|X}P_{T|X}.
\end{equation}
This assumption ensures that all relevant confounding factors are captured by the observed covariates, so that, after accounting for these covariates, the treatment assignment is independent of potential outcomes and can be treated as randomized. A graphical representation via a Bayesian network is provided in Appendix~\ref{appendix:scm_figure}.

Given $n$ units and the corresponding observed dataset $\mathcal{D} = \left\{ (X_i, Y^\text{obs}_i = Y_i(T_i), T_i)\right\}_{i = 1}^n$, the objective of this work is to reliably estimate the counterfactual outcome for a new unit by constructing prediction sets with marginal coverage guarantees.   Without loss of generality, consider a test unit with treatment assignment $T = 0$. For this unit we observe $\big(X, Y(0), T=0\big)$, and the target is the counterfactual outcome $Y^{\text{cf}} = Y(1)$. Our goal is to construct a prediction set $\Gamma(X)$ for $Y(1)$ satisfying the inequality
\begin{equation}
\label{eq:marginal_guarantee}
    \Pr\big(Y(1)\in \Gamma(X)\big) \geq 1- \alpha
\end{equation}
for some user-specified level $1-\alpha$. By (\ref{eq:marginal_guarantee}), the estimation set $\Gamma(X)$ covers the true counterfactual $Y(1)$ with probability no smaller than $1-\alpha$.

As in \citep{lei2021conformal}, to construct the set predictor $\Gamma(X)$, we assume access to pre-trained quantile regressors $\hat{q}^0_\gamma(X)$ and $\hat{q}^1_\gamma(X)$, which provide, respectively, estimates of the $\gamma$-quantiles, with $\gamma \in [0,1]$, for the potential outcomes $Y(0)$ and $Y(1)$ associated with the covariates $X$. No assumption is made on the accuracy of these estimators, which can be designed using techniques such as quantile random forests \citep{meinshausen2006quantile}, gradient-boosted quantile regression \citep{friedman2001greedy}, or abductive inference via structural causal models (SCMs) \citep{pearl2009causality, louizos2017causal}.

\section{Background: Conformal Inference for Counterfactual Outcomes}
\label{sec:background}

To construct valid estimation sets $\Gamma(X)$ for unobserved counterfactuals $Y^\text{cf} = Y(1)$, reference \citep{lei2021conformal} proposed a method based on weighted conformal prediction (WCP). The method, referred to here as \textit{conformal counterfactual inference} (CCI) leverages the pre-trained quantile regressor $\hat{q}^1_\gamma (X)$ of the counterfactual outcome $Y(1)$ given the covariates $X$. Henceforth, we use the simplified notation $\hat{q}_\gamma (X) = \hat{q}^1_\gamma (X)$. Furthermore, reference \citep{lei2021conformal} assumes the \textit{propensity score}
\begin{equation}
e(x) = \mathrm{Pr}(T = 1 \mid X = x),
\end{equation}
i.e., the probability (obtained from the joint distribution (\ref{eq:joint})) of assigning treatment variable $T = 1$ to a unit with covariates $X$, to be known. Additionally, it is assumed that the condition $0<e(x)<1$, known as the \textit{overlap condition}, holds almost surely. Under these conditions, the following steps are applied:

 \textbf{1. Split the calibration set by treatment:} Partition the calibration dataset $\mathcal{D} = \left\{(X_i,  Y_i(T_i), T_i)\right\}_{i = 1}^n$ into the treatment-specific datasets
   \begin{equation}
\label{eq:calibration_data}
\begin{split}
\mathcal{D}_0 &= \{(X_i, Y_i(0)) : T_i = 0\}\;\;\text{and} \\
\mathcal{D}_1 &= \{(X_i, Y_i(1)) : T_i = 1\},
\end{split}
\end{equation}
    with sizes $n_0$ and $n_1$, respectively, satisfying the equality $n = n_0 + n_1$.

\textbf{2. Compute the calibration nonconformity scores and importance weights:} For each point \( (X_i,  Y_i(1))\) in the calibration dataset $\mathcal{D}_1$, using the pre-trained quantile regressor, compute the estimated quantiles $\hat{q}_{\alpha_\text{lo}}(X_i)$ and $\hat{q}_{\alpha_\text{hi}}(X_i)$ for the outcome $Y_i(1)$, where the probabilities $\alpha_\text{lo}$ and $\alpha_\text{hi}$ are selected to satisfy the equality $1-\alpha = \alpha_\text{hi} - \alpha_\text{lo}$. The \emph{nonconformity score} is the standard conformalized quantile regression (CQR) score \citep{romano2019conformalized}:
    \begin{equation}
    \label{eq:nonconform_score}
        S_i = \max\left\{ \hat{q}_{\alpha_{\text{lo}}}(X_i) - Y_i(1),\; Y_i(1)- \hat{q}_{\alpha_{\text{hi}}}(X_i) \right\}.
    \end{equation}
    The score (\ref{eq:nonconform_score}) is negative when the outcome $Y_i(1)$ falls inside the estimated interval $[\hat{q}_{\alpha_\text{lo}}(X_i), \hat{q}_{\alpha_\text{hi}}(X_i)]$, and increases as the observation gets further away from the estimated interval bounds. For each data point \( (X_i, Y_i(1)) \in \mathcal{D}_1 \), evaluate also the importance weight $w_i = 1/e(X_i)$.

    \textbf{3. Evaluate the estimation set:} For a given test point \( X \) with $T = 0$, produce the estimation interval
    \begin{equation}
    \label{eq:interval}
        \Gamma(X) = \left[ \hat{q}_{\alpha_{\text{lo}}}(X) - \eta,\; \hat{q}_{\alpha_{\text{hi}}}(X) + \eta \right],
    \end{equation}
    where the interval widening parameter \( \eta \) is computed as
    \begin{equation}
    \label{eq:quantile_weights}
\eta(X) = \inf \left\{ t \in \mathbb{R} : \frac{\sum_{i=1}^{n_1}\mathbf{1}(S_i\leq t) w_i}{\sum_{i=1}^{n_1} w_i + \frac{1}{e(X)}} \geq 1 - \alpha \right\}.
    \end{equation}
    This selects the smallest value of $\eta$ such that the empirical weighted coverage over the real calibration points, adjusted for the test point, reaches the target level $1-\alpha$, thereby controlling the miscoverage rate at the desired level.

Reference \citep{lei2021conformal} proves that the marginal guarantee (\ref{eq:marginal_guarantee}) is satisfied by the estimation set (\ref{eq:interval}) regardless of the accuracy of the quantile model $\hat{q}_\gamma(X)$. More precisely, the condition (\ref{eq:marginal_guarantee}) is met by evaluating the probability over the joint distribution (\ref{eq:joint}) of the calibration data $\mathcal{D}$ used to compute the nonconformity scores and interval widening parameter $\eta$, as well as over the distribution of the test data point $(X, T, Y(0))$ for which the estimation interval is constructed.

\section{Efficient Conformal Inference with Synthetic Counterfactuals}
\label{sec:method}

The CCI approach reviewed in the previous section faces a key practical challenge \citep{shalit2017estimating}: there is often a significant imbalance in treatment assignment within observational datasets (see Fig. \ref{fig:data}). In particular, the dataset $\mathcal{D}_0$ encompassing data for untreated units can be much larger than the dataset $\mathcal{D}_1$ for the treated units, i.e., $n_0 \gg n_1$. For instance, in many medical applications, the number of treated units, i.e., with $T = 1$, is significantly smaller than the number of untreated ones, i.e., with $T = 0$. In fact, treatments are often costly or time-consuming to administer, while control data can be passively collected from existing records \citep{ballinari2024calibrating}.

Given a test unit with treatment variable $T = 0$, the state-of-the-art CCI method constructs estimation intervals (\ref{eq:interval}) for the counterfactual $Y(1)$ using real treated data $\mathcal{D}_1 = \{(X_i, Y_i): T_i = 1\}$ as calibration data. When the dataset $\mathcal{D}_1$ is small, the resulting intervals may become too wide to be useful. To address this limitation, in this section we introduce \textit{synthetic data-powered CCI} (SP-CCI), which augments the calibration dataset $\mathcal{D}_1$ with synthetic samples $\tilde{\mathcal{D}}_1$ whose counterfactual labels are generated from the covariates of the larger control set $\mathcal{D}_0$ (see Fig. \ref{fig:data}). We use this augmented dataset to calibrate estimation intervals for the counterfactual outcome $Y^\text{cf} = Y(1)$ that provide high-probability guarantees (\ref{eq:marginal_guarantee}) on the miscoverage rate.

\subsection{Generating Synthetic Counterfactual Labels}

To augment the calibration dataset $\mathcal{D}_1$ with synthetic data points generated from the dataset $\mathcal{D}_0$, we assume the availability of any pre-trained predictive model $\hat{P}_{Y(1)|X}$ for the outcome $Y(1)$ given the covariates $X$. The predictor $\hat{P}_{Y(1)|X}$ may be potentially implemented via a general-purpose existing model like a large language model (LLM) suitably prompted with a description of the task. We make no assumption on the quality of the model $\hat{P}_{Y(1)|X}$. By sampling from the model \( \hat{P}_{Y(1)|X}\), we obtain the counterfactual label
\begin{equation}
\hat{Y}_1(X) \sim \hat{P}_{Y(1)|X}.
\end{equation}

We also assume that the marginal treatment probability $P_T$ is known or that it can be estimated from data, with the effect of an inaccurate estimation studied in Sec. \ref{sec:theoretical_guarantees}.

Using the synthetic counterfactuals, the synthetic calibration dataset $\tilde{\mathcal{D}}_1$ is created as
\begin{equation}
\label{eq:synthetic_dataset}
\tilde{\mathcal{D}}_1 = \{ (\tilde{X}_i, \tilde{Y}_i) = (X_i, \hat{Y}_i(1))\}_{i = 1}^{n_0},
\end{equation}
where $X_i$ represents the covariates for the $i$-th data point of dataset $\mathcal{D}_0$. As shown in Fig. \ref{fig:data}, the dataset \( \tilde{\mathcal{D}}_1 \) is thus derived from the factual dataset \( \mathcal{D}_0 \) by assigning counterfactual labels to the covariates in dataset $\mathcal{D}_0$ using the predictive model $\hat{P}_{Y(1)\mid X}$.

While this process effectively increases the size of the calibration dataset available for the treatment arm, it introduces a new challenge towards guaranteeing the coverage condition (\ref{eq:marginal_guarantee}): the synthetic outcome \( \tilde{Y}_i \) in (\ref{eq:synthetic_dataset}) is only an approximation of the corresponding true counterfactual outcome $Y_i(1)$. Therefore, simply merging the datasets \( \mathcal{D}_1 \) and \( \tilde{\mathcal{D}}_1 \) and applying the method in \citep{lei2021conformal} to the resulting dataset would generally violate the coverage condition (\ref{eq:marginal_guarantee}).

\subsection{Constructing Reliable Estimation Sets using Synthetic Counterfactual Labels}

Given a test point \( (X, T = 0) \), in a manner similar to CCI (see (\ref{eq:interval})), we wish to construct an estimation interval \( \Gamma_\eta(X) = [\hat{q}_{\text{lo}}(X) - \eta,\; \hat{q}_{\text{hi}}(X) + \eta] \) for the counterfactual outcome \( Y(1) \), where \( \hat{q}_{\text{lo}}(X) \) and \( \hat{q}_{\text{hi}}(X) \) are the estimated lower and upper quantile for the counterfactual $Y(1)$ produced by the pre-trained model $\hat{q}_\gamma (X)$. Unlike CCI, the widening parameter \( \eta \) is calibrated to ensure that the coverage condition (\ref{eq:marginal_guarantee}) by leveraging not only the smaller dataset $\mathcal{D}_1$, but also larger synthetic dataset $\tilde{\mathcal{D}}_1$.

To this end, using both datasets $\mathcal{D}_1$ and $\tilde{\mathcal{D}}_1$, SP-CCI first obtains an unbiased estimate $\hat{L}_\eta$ of the miscoverage probability $L_\eta = \mathrm{Pr}\left(Y(1) \notin \Gamma_\eta(X)\right)$. Then, it evaluates an upper confidence bound $\hat{L}^+_\eta$ on the probability $L_\eta$ using the estimate $\hat{L}_\eta$. Finally, SP-CCI selects the parameter $\eta$ so that the upper confidence bound $\hat{L}^+_\eta$ does not exceed the target value $\alpha$. At a technical level, SP-CCI combines PPI \citep{angelopoulos2023prediction}, which is used to obtain the unbiased estimate $\hat{L}_\eta$, with RCPS \citep{bates2021distribution}, which supports the selection of the parameter $\eta$.

To elaborate, define the miscoverage loss for a given widening parameter $\eta$ and input-output pair $(X,Y)$ as
\begin{equation}
\label{eq:binary_miscoverage}
    \ell_\eta(X, Y) =
    \begin{cases}
        0 & \text{if } Y \in [\hat{q}_{\text{lo}}(X) - \eta,\; \hat{q}_{\text{hi}}(X) + \eta], \\
        1 & \text{otherwise}.
    \end{cases}
\end{equation}
The expectation of the loss (\ref{eq:binary_miscoverage}) with respect to the distribution of the variables $\left(X, Y(1)\right)$ is given by the miscoverage probability
\begin{equation}
\label{eq:expected_loss}
\begin{split}
L_\eta
&= \mathbb{E}\!\left[\ell_\eta(X, Y(1))\right]
= \Pr\!\big( Y(1) \notin \Gamma_\eta(X) \big),
\end{split}
\end{equation}
which we wish to control according to the inequality in (\ref{eq:marginal_guarantee}).

As mentioned, SP-CCI builds on an unbiased estimate $\hat{L}_\eta$ of the expected loss (\ref{eq:expected_loss}) that incorporates both real and synthetic calibration sets. To construct this estimate, as illustrated in Fig.~\ref{fig:grouping} (see Appendix~\ref{appendix:grouping_figure}), we partition the $n_0 > n_1$ examples in the dataset $\tilde{\mathcal{D}}_1$ into $n_1$ groups $\{\tilde{\mathcal{D}}_{1, i}\}_{i = 1}^{n_1}$ of $r =  \lfloor n_0 / n_1 \rfloor$ data points each \citep{einbinder2024semi, park2025adaptive}. Each group $\tilde{\mathcal{D}}_{1, i} = \{(\tilde{X}_j, \tilde{Y}_j)\}_{j = r(i-1)+1}^{ri}$ is assigned to a different real calibration point \( (X_i, Y_i) \in \mathcal{D}_1 \).

Furthermore, SP-CCI computes the modified weights
\begin{equation}
\label{eq:new_weights}
w_i = \frac{P_{X_i}( X_i)}{P_{X_i|T}(X_i \mid 1)} = \frac{P_T( 1)}{e(X_i)}
\end{equation}
for all $n_1$ real data points in dataset $\mathcal{D}_1$, and
\begin{equation}
\label{eq:new_weights_2}
\tilde{w}_i = \frac{P_{X_i}( \tilde{X}_i)}{P_{X_i|T}(\tilde{X}_i \mid 0)} = \frac{P_T( 0)}{1-e(\tilde{X}_i)}
\end{equation}
for all $n_0$ synthetic data points in dataset $\tilde{\mathcal{D}}_1$. Note that, unlike \citep{lei2021conformal}, the evaluation of the weights (\ref{eq:new_weights})-(\ref{eq:new_weights_2}) requires knowledge not just of the propensity score $e(\cdot)$, but also of the treatment probability $P_T ( 1)$. The effect of a misspecified treatment probability $P_T ( 1)$ is studied in Sec. \ref{sec:theoretical_guarantees}.

Using the datasets $\mathcal{D}_1$ and $\tilde{\mathcal{D}}_1$, SP-CCI constructs an estimate of the miscoverage probability (\ref{eq:expected_loss}) given by
\begin{equation}
\label{eq:l_tilde}
    \hat{L}_\eta = \frac{1}{n_1} \sum_{i=1}^{n_1} \hat{\ell}_{i,\eta},
\end{equation}
where $\hat{\ell}_{i,\eta}$ is the estimate obtained using the $i$-th data point $(X_i, Y_i)$ from the real dataset $\mathcal{D}_1$, as well as the corresponding group $\tilde{\mathcal{D}}_{1, i}= \{(\tilde{X}_j, \tilde{Y}_j)\}_{j = r(i-1)+1}^{ri}$ from the synthetic dataset. Taking inspiration from PPI \citep{angelopoulos2023prediction, einbinder2024semi, park2025adaptive}, this estimate is obtained as
\begin{equation}
\label{eq:debiased_loss_eta}
\hat{\ell}_{i,\eta}
= \frac{1}{r} \sum_{j = r(i-1)+1}^{ri}
    \tilde{w}_j \,\ell_\eta(\tilde{X}_j, \tilde{Y}_j) - w_i \Big[ \ell_\eta(X_i, \hat{Y}_i) - \ell_\eta(X_i, Y_i) \Big],
\end{equation}
where $\hat{Y}_i\sim \hat{P}_{Y(1)|X}$ represents an estimate of the outcome $Y_i(1)$ corresponding to the covariate $X_i$ in dataset $\mathcal{D}_1$. An intuitive interpretation of the two terms in \eqref{eq:debiased_loss_eta} is provided in Appendix~\ref{appendix:proofs}.

The quantity \( \hat{L}_\eta \) in (\ref{eq:l_tilde}) can be shown to be an unbiased estimator of the expected loss $L_\eta$ (see Appendix~\ref{appendix:proofs}). Furthermore, by Hoeffding's inequality, due to the fact that the terms $\hat{\ell}_{i, \eta}$ are bounded in the interval $[-1/\min_x\{e(x)\},1/\min_x\{e(x)\} + 1/(1-\max_x\{e(x)\}) ]$ almost surely, we have the upper confidence bound on the miscoverage probability $L_\eta$ \citep{bates2021distribution}
\begin{equation}
\label{eq:Hoeffding}
    \mathrm{Pr} \left( L_\eta \leq \hat{L}_\eta^+ = \hat{L}_\eta + C\sqrt{\frac{1}{2n_1} \log\left(\frac{1}{\delta}\right)} \right) \geq 1 - \delta
\end{equation}
for any probability $\delta$, and $C =2/\min_x\{e(x)\} + 1/(1-\max_x\{e(x)\}) $.

The estimation interval is finally given by
\begin{equation}
\label{eq:interval_augmented}
\Gamma(X) = \Gamma_{\hat{\eta}}(X) = [\hat{q}_{\text{lo}}(X) - \hat{\eta},\; \hat{q}_{\text{hi}}(X) + \hat{\eta}],
\end{equation}
where the widening parameter $\hat{\eta}$ is selected so as to ensure that the upper bound $\hat{L}^+_\eta$ on the miscoverage probability is within the target level \( \alpha \), i.e., \citep{einbinder2024semi}
\begin{equation}
\label{eq:eta}
    \hat{\eta} = \min\left\{ \eta \geq 0 : \hat{L}^+_\eta \leq \alpha \right\}.
\end{equation}

\subsection{Theoretical Guarantees}
\label{sec:theoretical_guarantees}

In this section, we show that the proposed SP-CCI estimation set (\ref{eq:interval_augmented}) satisfies the marginal coverage requirement (\ref{eq:marginal_guarantee}) with probability no smaller than $1-\delta$. We first consider the case where the importance weights in (\ref{eq:new_weights}) are known exactly, and then we analyze the impact of a mismatch between the weights used in (\ref{eq:debiased_loss_eta}) and the true weights (\ref{eq:new_weights})-(\ref{eq:new_weights_2}). Note that mismatches in the weights may result from an imprecise knowledge of the treatment probability $p_T (1)$, even when the propensity score $e(\cdot)$ is known.

\textbf{Comparison of guarantee types:} SP-CCI's guarantee \eqref{eq:updated_condition} is a \emph{high-probability} guarantee over the draw of the calibration data, which is controlled by the user-specified parameter $\delta$. CCI's guarantee, by contrast, is an \emph{unconditional} marginal coverage guarantee, which holds \emph{on average} over the draw of the calibration data. Thus, via the choice of parameter $\delta$, SP-CCI provides a tunable outage control mechanism that the unconditional guarantee of CCI does not offer. A sensitivity analysis to $\delta$ is provided in Appendix~\ref{appendix:delta_sensitivity}.

\begin{proposition}
\label{prop:1}
    For any test point $(X,T = 0)$, and for any probability $0<\delta<1$, the SP-CCI estimation interval $\Gamma(X)$ in (\ref{eq:interval_augmented}) satisfies the condition
\begin{equation}
\label{eq:updated_condition}
    \mathrm{Pr}\left(\mathrm{Pr}\left(Y(1)\in \Gamma(X)\mid \mathcal{D}_1, \tilde{\mathcal{D}}_1\right)\geq 1- \alpha\right)\geq 1-\delta,
\end{equation}
where the inner probability is taken over the randomness of the test point $(X, T = 0, Y(1))$, while the outer probability is evaluated over the distribution of the calibration datasets $\mathcal{D}_1$ and $\tilde{\mathcal{D}}_1$ used to compute the estimation interval $\Gamma(X)$.
\end{proposition}

The next result shows that the coverage guarantee (\ref{eq:updated_condition}) can be retained even in the presence of a weight estimation error, as long as one suitably increases the widening parameter (\ref{eq:eta}) to account for the quality of the estimated importance weights.

\begin{proposition}
\label{prop:2}
Let $\hat{w}_i$ and $\hat{\tilde{w}}_i$ denote estimates of the weights $w_i$ and $\tilde{w}_i$, respectively, and assume that these estimates are used in lieu of the weights $w_i$ and $\tilde{w}_i$ in (\ref{eq:debiased_loss_eta}). Assume also that the estimated importance weights \( \hat{w}_i \) and \( \hat{\tilde{w}}_i \) satisfy the inequalities
\begin{equation}
\label{eq:estimated_weights}
 | \hat{w}_i - w_i | \leq \epsilon,\;\;\text{and}\;\; | \hat{\tilde{w}}_i - \tilde{w}_i | \leq \tilde{\epsilon}
\end{equation}
for all data points for some $\epsilon\geq 0$ and $\tilde{\epsilon}\geq 0$. Then, the SP-CCI estimation interval \( \Gamma(X) \) constructed using the estimated weights and with the widening parameter
\begin{equation}
\label{eq:new_choice}
\hat{\eta} = \min\left\{ \eta \geq 0 :\tilde{L}_\eta + \epsilon + \tilde{\epsilon}+ \sqrt{\frac{1}{2n_1} \log\left( \frac{1}{\delta} \right)} \leq \alpha\right\}
\end{equation}
satisfies the probabilistic guarantee (\ref{eq:updated_condition}), where $\tilde{L}_\eta$ is calculated as in (\ref{eq:l_tilde}) by using the estimated weights $\hat{w}_i$ and $\hat{\tilde{w}}_i$ from (\ref{eq:estimated_weights}).
\end{proposition}

Beyond coverage validity, SP-CCI admits theoretical efficiency guarantees.
In particular, an extension based on a weighted prediction-powered calibration
can be shown to yield prediction intervals that are asymptotically no wider
than those obtained using real treated data only.
Formal statements and proofs are provided in Appendix~\ref{appendix:efficiency_proof}.

\section{Experiments}
\label{sec:experiments}

In this section, we empirically validate the proposed SP-CCI method and compare it
against CCI \citep{lei2021conformal} and the GESPI framework \citep{bashari2026statistical} through experiments on a synthetic dataset
(Sec.~\ref{sec:synthetic_exp}) and a semi-synthetic dataset
(Sec.~\ref{sec:semi-synthetic}).
Additional experiments are reported in the appendix, including a real-world policy
evaluation with counterfactual loss (Appendix~\ref{sec:real-world}) and an experiment
using a truly pretrained counterfactual generator under extreme treatment imbalance
(Appendix~\ref{appendix:twins}). We evaluate performance primarily in terms of
efficiency, measured by the size of the predicted counterfactual prediction intervals.

\textbf{GESPI baseline:} We instantiate GESPI \citep{bashari2026statistical} with CCI \citep{lei2021conformal} as the base method and the same synthetic dataset $\tilde{\mathcal{D}}_1$. Following the GESPI framework, CCI is run three times: (1) on real data $\mathcal{D}_1$ at the target level $\alpha$; (2) on real data $\mathcal{D}_1$ at a relaxed guardrail level $\alpha + \varepsilon$; and (3) on the pooled real and synthetic data $\mathcal{D}_1 \cup \tilde{\mathcal{D}}_1$ at level $\alpha$. The outputs are aggregated by intersecting the three intervals. GESPI guarantees that the error rate never exceeds $\alpha + \varepsilon$ regardless of synthetic data quality, where $\varepsilon$ is a user-chosen parameter; we set $\alpha + \varepsilon  =0.15$ in our experiments to ensure a fair comparison between the benchmarks (an alternative setting is also considered in Sec. \ref{sec:semi-synthetic}). A detailed description of GESPI is provided in Appendix~\ref{appendix:gespi}.

\subsection{Efficiency Advantages on Synthetic Data}
\label{sec:synthetic_exp}

We begin by evaluating SP-CCI on the same simulation setup used in \citep{lei2021conformal}.

\paragraph{Data Generation:} We follow the simulation setup of \citep{lei2021conformal, wager2018estimation} exactly; full details are given in Appendix~\ref{appendix:synthetic_setup}. Briefly, we generate $n=5000$ samples with 10-dimensional covariates, a beta propensity model, and a nonlinear treated outcome. We use a 30/30/20/20 split for quantile training, generator training, calibration, and testing, and consider LQ/MQ/HQ generators trained on 20/60/100\% of the generator training data. We set $\alpha = 0.15$ and $\delta = 0.1$, following the experimental setup of \citep{einbinder2024semi}.

\paragraph{Implementation:} The quantile regressor $\hat{q}_\gamma(\cdot)$ is implemented as two separate gradient-boosted regression models \citep{friedman2001greedy} for $\gamma = \alpha/2$ and $\gamma = 1 - \alpha/2$, using treated data points ($T = 1$) from dataset \(\mathcal{D}_{\hat{q}}\) reserved for quantile estimation. We adopt the quantile loss, a learning rate of 0.1, and 500 boosting stages. The generative model $\hat{P}_{Y(1)\mid X}$ is implemented as a neural network regression model trained on treated units ($T=1$), predicting $Y(1)$ from covariates $X$.

\begin{figure}[t]
    \centering
    \begin{subfigure}[t]{0.48\linewidth}
        \centering
        \includegraphics[width=\linewidth]{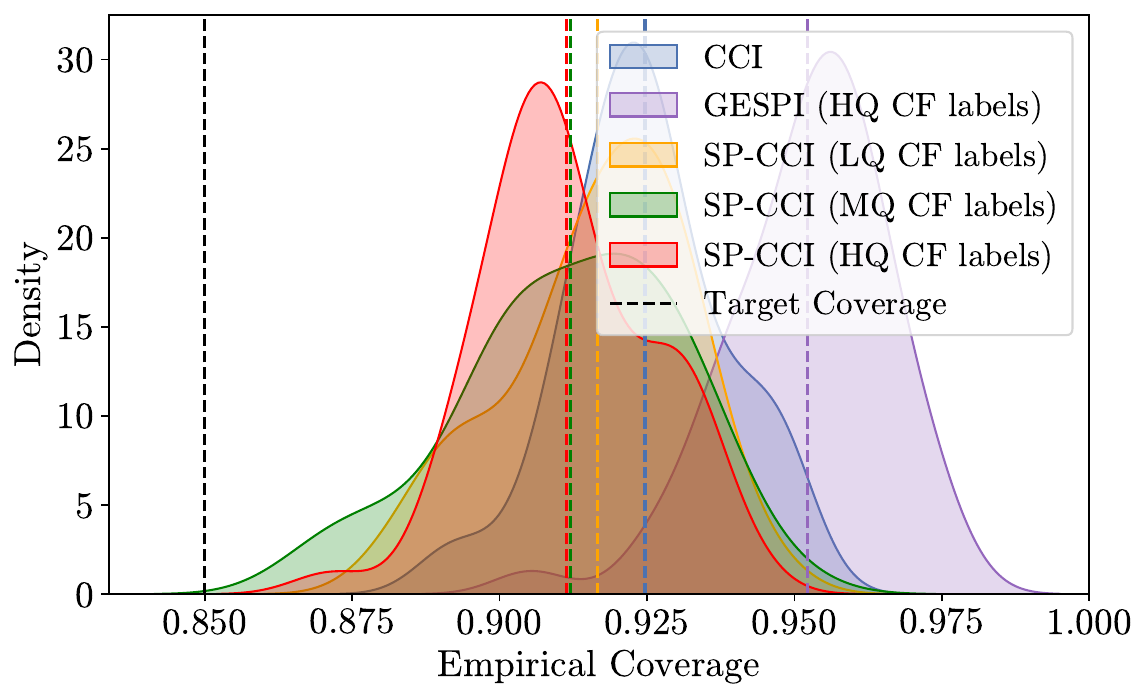}
        \caption{}
        \label{fig:coverage}
    \end{subfigure}
    \hfill
    \begin{subfigure}[t]{0.48\linewidth}
        \centering
        \includegraphics[width=\linewidth]{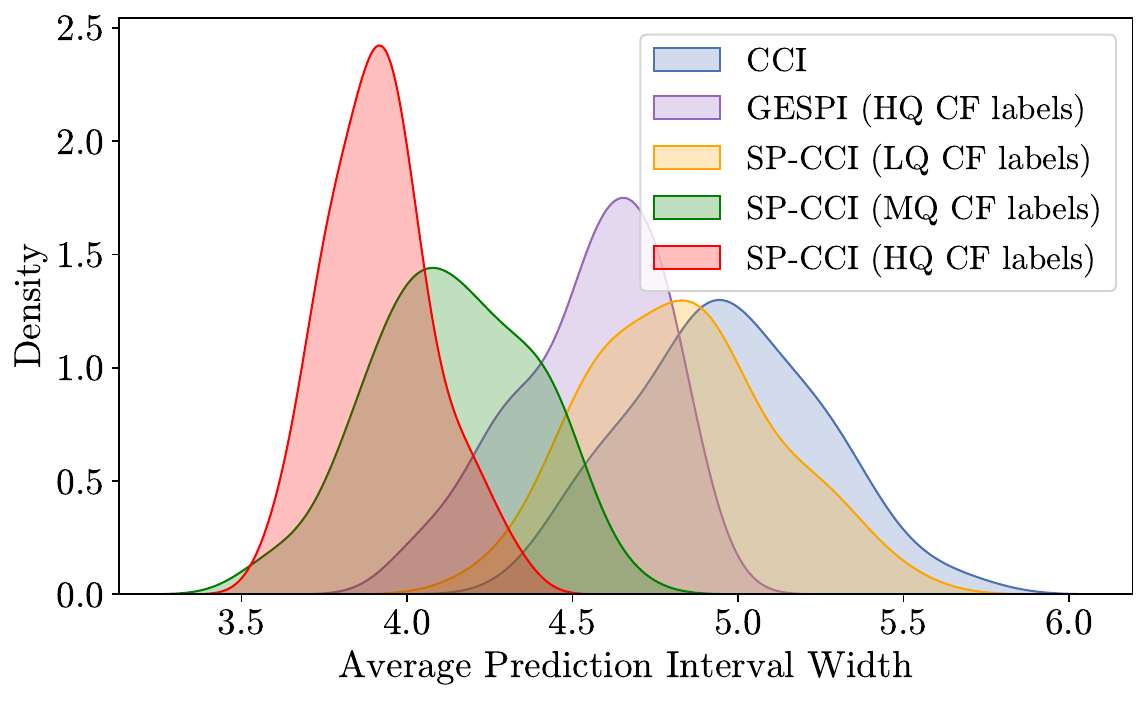}
        \caption{}
        \label{fig:width_eta}
    \end{subfigure}
    \caption{
    Synthetic data example from \citep{lei2021conformal}: (a) Distribution of empirical test coverage for CCI \citep{lei2021conformal}, GESPI \citep{bashari2026statistical}, and SP-CCI (with counterfactual labels of different quality levels) evaluated over 50 independent realizations of the data. The black dashed line indicates the target level $1 - \alpha = 0.85$, while the other dashed lines represent the average empirical test coverage probabilities.
    (b) Distribution of the average test prediction interval width. (LQ/MQ/HQ: low-/medium-/high-quality; CF: counterfactual)
    }
    \label{fig:coverage_vs_width}
\end{figure}

\paragraph{Results and discussion:} Fig.~\ref{fig:coverage} shows the distribution of the empirical test marginal coverage rates for CCI, GESPI, and SP-CCI across 50 runs over different random splits of the available dataset. By their respective theoretical properties, CCI meets the nominal coverage level of $1 - \alpha = 0.85$ on average (dashed lines), while all SP-CCI variants and GESPI meet this coverage level with a probability higher than $1-\delta = 0.9$. Note that SP-CCI's (coverage violation rate) CVR of up to 6\% across runs is a direct consequence of its high-probability guarantee type (see Sec.~\ref{sec:theoretical_guarantees}); this is expected behavior, not a failure of coverage control. Fig.~\ref{fig:width_eta} presents the corresponding distribution of the average test prediction interval width. The results demonstrate that SP-CCI consistently achieves narrower intervals compared to both CCI and GESPI, while still satisfying the coverage guarantee in (\ref{eq:updated_condition}). Furthermore, as the quality of the counterfactual generative model improves, from LQ to HQ, the interval width decreases, confirming that higher-quality synthetic data yields tighter and more informative prediction intervals.

\subsection{Real-World Validation on the IHDP Dataset}
\label{sec:semi-synthetic}

We next validate SP-CCI on a semi-synthetic benchmark derived from the Infant Health and Development Program (IHDP), a widely used testbed for counterfactual inference \citep{hill2011bayesian}. In this dataset, covariates \( X \in \mathbb{R}^{25} \) represent real-world demographic and health-related attributes of premature infants and their mothers, such as birth weight, gestational age, and maternal education level. The potential outcomes \( Y(0) \) and \( Y(1) \) denote simulated measures of cognitive development under control and treatment, respectively, with the treatment corresponding to participation in an early childhood intervention program. In the original study, treatment was assigned at random, but reference \citep{hill2011bayesian} introduced selection bias by removing a non-random subset of treated units. This created a treatment--control imbalance, resulting in a dataset with a treated-to-control ratio of approximately one to four.

We use the original IHDP train/test splits: all model training and calibration are performed exclusively on the training set (747 samples), while the test set is reserved solely for final evaluation. This strict separation ensures no train-test leakage, in line with standard causal evaluation practice.

\paragraph{Quantile estimation and generative models:} We use a DSCM \citep{pawlowski2020deep} to estimate counterfactual quantiles $\hat{q}_\gamma(X)$ via abduction, action, and prediction using 100 Monte Carlo samples; full implementation details are given in Appendix~\ref{appendix:ihdp_setup}. Since the DSCM requires knowledge of $T$, it cannot serve directly as the SP-CCI generative model. We therefore consider two generative models $\hat{P}_{Y(1)\mid X}$: (1) a neural network regression model trained on treated units in-domain, as in Sec.~\ref{sec:synthetic_exp}; and (2) a pre-trained LLaMA-2-13B LLM, with covariates formatted as a key-value text record.

\paragraph{Results and Discussion:}
\begin{wraptable}{r}{0.52\columnwidth}
\vspace{-0.8em}
\centering
\small
\resizebox{0.52\columnwidth}{!}{%
\begin{tabular}{lcc}
\toprule
\textbf{Method} & \textbf{APIW} & \textbf{CVR} \\
\midrule
CCI \citep{lei2021conformal}                                              & 20.24         & 3\% \\
GESPI (HQ, $\alpha{+}\varepsilon{=}0.15$) \citep{bashari2026statistical} & 5.562         & 4\% \\
GESPI (HQ, $\alpha = 0.15, \;\varepsilon{=}0.05$)                                & 3.491 & 42\%            \\
SP-CCI (LQ)                                                               & 4.308         & 6\%             \\
SP-CCI (MQ)                                                               & 4.022         & 4\%             \\
SP-CCI (HQ)                                                               & 3.521         & 2\%             \\
SP-CCI (LLM)                                                              & 3.893         & 3\%             \\
\bottomrule
\end{tabular}
}
\caption{APIW and CVR on IHDP (50 runs, $\alpha=0.15$). For GESPI, two variants are shown: a lenient guardrail ($\alpha{+}\varepsilon{=}0.15$) and a tight guardrail (small $\varepsilon$). (LQ/MQ/HQ: label quality; LLM: LLaMA-2-13B)}
\label{tab:ihdp_width_results}
\vspace{-0.5em}
\end{wraptable}

Table~\ref{tab:ihdp_width_results} reports the average prediction interval width (APIW) and CVR over 50 independent runs on IHDP. When GESPI is tightened to a small $\varepsilon$, it produces narrower intervals than SP-CCI, but at the cost of a CVR of 42\%, a level that is untenable in high-stakes settings such as the medical decision-making scenario motivating this work. SP-CCI, by contrast, exposes parameter $\delta$ as an explicit outage knob, keeping CVR between 2\% and 6\% across all variants, while consistently outperforming CCI and GESPI with the guardrail $\alpha + \epsilon = 0.15$.

Notably, SP-CCI with the pretrained LLM-based generator achieves performance competitive with high-quality in-domain simulators, despite the LLM being trained entirely outside the IHDP dataset, demonstrating that SP-CCI can effectively leverage externally available pretrained models under distribution shift.

\section{Conclusion and Future Work}
\label{sec:conclusion}

In this paper, we introduced SP-CCI, a synthetic data-powered extension of conformal counterfactual inference designed to address the efficiency limitations of CCI in imbalanced treatment settings. By augmenting the calibration set with synthetic counterfactual labels and applying a debiased miscoverage estimator inspired by PPI, SP-CCI achieves high-probability marginal coverage guarantees while producing substantially narrower prediction intervals. Theoretical analysis establishes robustness to importance weight misspecification, and experiments on various datasets demonstrate consistent efficiency gains over CCI and GESPI.

Future work includes extending SP-CCI to also leverage synthetic data for calibrating the control group; the possibility of using counterfactual generative models conditioned on both covariates $X$ and treatment $T$ to better capture treatment--covariate interactions; extending the framework to multi-arm and continuous treatments; and developing tighter variance-adaptive concentration bounds.

\section*{Acknowledgments}
The work of A. Farzaneh and O. Simeone was supported by the European Research Council (ERC) under the European Union’s Horizon Europe Programme (grant agreement No. 101198347). The work of O. Simeone was also supported by an EPSRC Open Fellowship (EP/W024101/1) and by the EPSRC project (EP/X011852/1).

\newpage
\appendix
\onecolumn

\section{Bayesian Network for Observational Setup}
\label{appendix:scm_figure}

\begin{figure}[h]
    \centering
    \includegraphics[width = 0.5\linewidth]{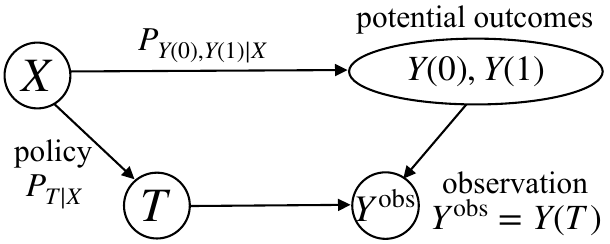}
    \caption{A Bayesian network representation of the observational setup for the potential outcomes framework under the SUTVA assumption and the strong ignorability assumption (\ref{eq:ignorability}). The covariates \( X \) are correlated with the treatment through the assigned policy \( T \sim P_{T\mid X} \) and also with the potential outcomes \( (Y(0), Y(1)) \sim P_{Y(0), Y(1)\mid X} \), with the observed outcome given by \( Y^{\text{obs}} = Y(T) \). By the assumption (\ref{eq:ignorability}), the treatment $T$ is correlated with the potential outcomes $(Y(0), Y(1))$ only through the covariates $X$.}
    \label{fig:scm_observational}
\end{figure}

\section{SP-CCI Grouping Scheme}
\label{appendix:grouping_figure}

\begin{figure}[h]
    \centering
    \includegraphics[width=0.8\linewidth]{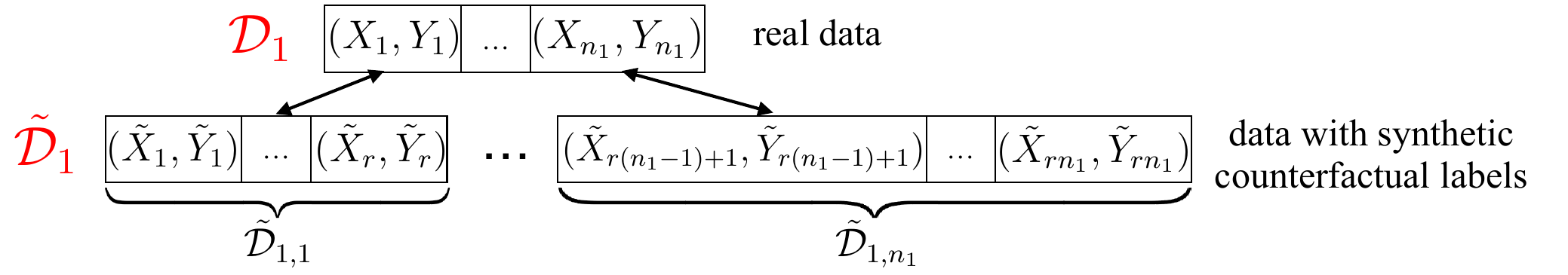}
    \caption{SP-CCI partitions the synthetic dataset $\tilde{\mathcal{D}}_1$ into $n_1$ disjoint groups $\{\tilde{\mathcal{D}}_{1,i}\}_{i = 1}^{n_1}$, each with $r$ data points. Each group $\tilde{\mathcal{D}}_{1,i}$ is assigned to a real data point $(X_i, Y_i)$ from the dataset $\mathcal{D}_1$.}
    \label{fig:grouping}
\end{figure}

\section{Further Related Work}
\label{appendix:related_work}

\noindent \textbf{Counterfactual inference:} Counterfactuals and treatment effects can be estimated using meta-learners \citep{kunzel2019metalearners}, representation learning \citep{johansson2016learning}, and architectures such as CFR/TARNet \citep{shalit2017estimating}, as well as Bayesian and tree-based approaches like BART \citep{hill2011bayesian} and causal forests \citep{wager2018estimation}. Generative models (e.g., CEVAE \citep{louizos2017causal}) and structural causal models \citep{pawlowski2020deep} learn latent structure for counterfactual estimation. All such methods provide point predictions rather than finite-sample, distribution-free intervals with coverage guarantees.

\noindent \textbf{Conformal prediction for counterfactuals:} Conformal inference has been adapted for counterfactuals in several ways: CCI \citep{lei2021conformal}, conformal sensitivity analysis \citep{yin2024conformal}, and conformal meta-learners for ITEs \citep{alaa2023conformal}. Despite differences in conformity scores and estimands, they all suffer in imbalanced settings, where calibration data scarcity yields wide intervals.

\noindent \textbf{Synthetic data in causal inference:} Generative models such as CEVAE \citep{louizos2017causal}, GANITE \citep{yoon2018ganite}, and SCIGAN \citep{bica2020estimating} impute missing counterfactuals, while model-based off-policy methods generate unobserved rewards for evaluation \citep{swaminathan2015counterfactual, thomas2016data}. These approaches reduce variance but risk bias in conformal settings. Semi-supervised risk control via PPI \citep{einbinder2024semi} addresses bias correction by calibrating with model predictions. The proposed SP-CCI applies PPI to synthetic counterfactuals to provide statistical guarantees on counterfactual estimation.

\section{Proofs}
\label{appendix:proofs}

\textbf{Interpretation of the debiased estimator:} The estimator \eqref{eq:debiased_loss_eta} combines a (weighted) empirical estimate from synthetic data with a correction term derived from real data. Specifically, the first term in (\ref{eq:debiased_loss_eta}) averages the miscoverage loss over the $r$ synthetic samples in group $\tilde{\mathcal{D}}_{1, i}$, which are scaled by their importance weights $\tilde{w}_j$. The second term adjusts for the potential bias in the synthetic counterfactual labels $\tilde{Y}_j$ by subtracting an estimate of the bias, computed as the difference between the loss on the true outcome $Y_i$ and its synthetic estimate $\hat{Y}_i$, scaled by weight $w_i$.

\begin{proposition}
    The quantity \( \hat{L}_\eta \) defined in (\ref{eq:l_tilde}) is an unbiased estimator of the expected loss \( L_\eta \) in (\ref{eq:expected_loss}).
\end{proposition}
\begin{proof}
    Recall from (\ref{eq:l_tilde}) and (\ref{eq:debiased_loss_eta}) that
    \[
    \hat{L}_\eta = \frac{1}{n_1} \sum_{i=1}^{n_1} \left( \frac{1}{r} \sum_{j=r(i-1)+1}^{ri} \tilde{w}_j \ell_\eta(\tilde{X}_j, \tilde{Y}_j)
    - w_i \left[\ell_\eta(X_i, \hat{Y}_i) - \ell_\eta(X_i, Y_i)\right] \right).
    \]
    Taking the expectation over all data-generating randomness, and using linearity of expectation, we obtain
    \begin{align}
        \mathbb{E}[\hat{L}_\eta] &= \frac{1}{n_1} \sum_{i=1}^{n_1} \left( \frac{1}{r} \sum_{j} \mathbb{E}\left[\tilde{w}_j \ell_\eta(\tilde{X}_j, \tilde{Y}_j)\right]
        - \mathbb{E}\left[w_i \ell_\eta(X_i, \hat{Y}_i)\right]
        + \mathbb{E}\left[w_i \ell_\eta(X_i, Y_i)\right] \right). \label{eq:exp_hatL}
    \end{align}
    Now, by the definitions of the importance weights in (\ref{eq:new_weights})--(\ref{eq:new_weights_2}), we have the identities
    \begin{align}
        \mathbb{E}\left[\tilde{w}_j \ell_\eta(\tilde{X}_j, \tilde{Y}_j)\right] &= \mathbb{E}_{p(X, \hat{Y}(1))} \left[\ell_\eta(X, \hat{Y})\right], \\
        \mathbb{E}\left[w_i \ell_\eta(X_i, \hat{Y}_i)\right] &= \mathbb{E}_{ p(X, \hat{Y}(1))} \left[\ell_\eta(X, \hat{Y})\right], \\
        \mathbb{E}\left[w_i \ell_\eta(X_i, Y_i)\right] &= \mathbb{E}_{p(X, Y(1))} \left[\ell_\eta(X, Y(1))\right] = L_\eta.
    \end{align}
    Substituting into (\ref{eq:exp_hatL}), and noting that the first two terms cancel exactly: $\mathbb{E}[\hat{L}_\eta] = L_\eta$. \qedhere
\end{proof}

\begin{proof}[Proof of Proposition~\ref{prop:1}]
Combining (\ref{eq:Hoeffding}) and (\ref{eq:eta}), and given (\ref{eq:expected_loss}), we have
\begin{equation}
    \mathrm{Pr} \left( \mathrm{Pr}\left( Y(1) \notin [\hat{q}_{\text{lo}}(X) - \eta,\; \hat{q}_{\text{hi}}(X) + \eta]  \right) \leq \alpha \right) \geq 1 - \delta,
\end{equation}
which is the same as condition (\ref{eq:updated_condition}).
\end{proof}

\begin{proof}[Proof of Proposition~\ref{prop:2}]
    Given that all the terms in (\ref{eq:debiased_loss_eta}) by which the weights get multiplied are in the interval $[-1, 1]$, we obtain
\begin{equation}
\left| \mathbb{E}[\tilde{L}_\eta] - \bar{\ell}_\eta \right|
\leq \epsilon + \tilde{\epsilon},
\end{equation}
where $\tilde{L}_\eta$ is calculated as in (\ref{eq:l_tilde}) by using the estimated weights $\hat{w}_i$ and $\hat{\tilde{w}}_i$ from (\ref{eq:estimated_weights}). Consequently, the guarantee (\ref{eq:updated_condition}) still holds as long as the widening parameter \( \hat{\eta} \) is chosen as per (\ref{eq:new_choice}).
\end{proof}

\section{Efficiency Guarantees and Width Scaling}
\label{appendix:efficiency_proof}

In this section, we propose and analyze SP-CCI++, a generalization of SP-CCI that mitigates potential inefficiencies due to imprecise synthetic labels by leveraging the PPI++ framework \citep{angelopoulos2023ppi++}. We show that SP-CCI++ retains the same coverage guarantees as SP-CCI, while yielding interval predictors that are asymptotically no wider than those obtained using real data only.

Following the PPI++ framework \citep{angelopoulos2023ppi++}, we define a weighted generalization of the debiased miscoverage estimator in \eqref{eq:debiased_loss_eta} that uses a weighting parameter $\lambda \in [0,1]$ to control the contribution of synthetic data as
\begin{equation}
\label{eq:debiased_loss_eta_pp}
\hat{\ell}_{i,\eta}(\lambda)
=
\frac{1}{r}
\sum_{j=r(i-1)+1}^{ri}
\lambda\,\tilde{w}_j\,\ell_\eta(\tilde{X}_j,\tilde{Y}_j)
-
w_i
\left[
\lambda\,\ell_\eta(X_i,\hat{Y}_i)
-
\ell_\eta(X_i,Y_i)
\right].
\end{equation}
The corresponding estimator of the miscoverage probability is given by $\hat{L}_\eta(\lambda) = \frac{1}{n_1}\sum_{i=1}^{n_1}\hat{\ell}_{i,\eta}(\lambda)$.
Note that for a parameter $\lambda = 0$ the estimator recovers real data only calibration based on RCPS \citep{bates2021distribution}, while for $\lambda = 1$ it recovers the SP-CCI estimator from Sec.~\ref{sec:method}.

To maximize the efficiency of SP-CCI++, we adopt the framework of \citealt{park2025adaptive} to tune the parameter $\lambda$ based on the quality of the synthetic labels. Following \citealt{park2025adaptive}, for any fixed $\lambda$, the loss terms $\{\hat{\ell}_{i,\eta}(\lambda)\}^{n_1}_{i=1}$ are used to construct an e-process $\{E^\lambda_i(\eta)\}^n_{i=1}$ for testing the null hypothesis $H_{0,\eta}:\mathbb{E}[\ell_{\eta}(X,Y(1)]\geq\alpha$. A valid e-process is given by the canonical exponential process \citep{waudby2024estimating}
\begin{align}
    \label{eq:e-process_exp}
    E^\lambda_i(\eta)=\prod^i_{t=1}\exp\left(\sqrt{\frac{8\log(1/\delta)}{n_1}}(\alpha-\hat{\ell}_{i,\eta}(\lambda))-\frac{\log(1/\delta)}{n_1}\right).
\end{align}
Fix a discretization $\Lambda\subset[0,1]$ including $\lambda=0$, and a prior $p(\lambda)$ with $p(0)>0$. SP-CCI++ sets its widening parameter based on the mixture e-process
$E_i(\eta)=\sum_{\lambda\in\Lambda}p(\lambda)E^\lambda_i(\eta)$ as
\begin{align}
    \label{eq:scpi++}
    \hat{\eta}^{++}=\min\left\{\eta\geq0: E_{n_1}(\eta')\geq 1/\delta, \forall \eta'>\eta\right\}.
\end{align}
We denote as $\Gamma^{++}(X)$ the resulting SP-CCI++ interval predictor.

\begin{theorem}
\label{thm:appendix_no_underperform}
 For any test point $(X,T = 0)$, and for any probability $0<\delta<1$, the SP-CCI++ estimation interval  $\Gamma^{++}(X)$  satisfies the condition (\ref{eq:updated_condition}).
Moreover, denoting by $\hat{\eta}(0)$ the widening parameter obtained using $\lambda = 0$ (real data only calibration), the following limit holds
\begin{equation}
\label{eq:no_underperform_limit}
\lim_{\delta \to 0,\; n_1 \to \infty}
\Pr\!\left(
\hat{\eta}^{++} \le \hat{\eta}(0)
\right)
=
1.
\end{equation}
Consequently, SP-CCI++ is asymptotically no wider than prediction intervals obtained using real treated data only.
\end{theorem}

\begin{proof}
\emph{Coverage guarantee:}
Fix any $\eta \ge 0$. The mixture e-process $E_i(\eta)$ is a convex combination of valid e-processes, hence itself a valid e-process under $H_{0,\eta}$. By Ville's inequality,
$\Pr\left(\sup_{i \le n_1} E_i(\eta) \ge 1/\delta \mid H_{0,\eta}\right)\le \delta$,
so RCPS with this e-process yields (\ref{eq:updated_condition}).

\emph{Efficiency guarantee:} Pick any $\lambda\in\Lambda$.
Define the log-growth rate $g_\lambda(\eta) = \lim_{n\to\infty}\frac{1}{n}\log E^{\lambda}_n(\eta)$,
$\eta_\lambda^\star = \inf\{ \eta : g_\lambda(\eta) > 0 \}$,
and $g_\lambda^\star = g_\lambda(\eta_\lambda^\star)$.
Let $n_\lambda' = \lceil(1+\epsilon)(\log(1/\delta)+\log(1/p(\lambda)))/g_\lambda^\star\rceil$ for any $\epsilon\in(0,1)$.
Since $n_\lambda'\to\infty$ as $\delta\to 0$, by a standard law-of-large-numbers argument for e-processes,
$\Pr(\frac{1}{n_\lambda'}\log E_{n_\lambda'}^\lambda(\eta) < g_\lambda(\eta)/(1+\epsilon))\to 0$.
Using the universal portfolio bound $\log E_n(\eta)\ge \log p(\lambda)+\log E_n^\lambda(\eta)$ \citep{cover1991universal},
and a union bound over a finite grid, $\Pr(\hat{\eta}^{++}\le \eta_\lambda^\star)\ge 1-N\xi_\lambda(\delta)$ where $\xi_\lambda(\delta)\to 0$.
Taking $\lambda=0$, combining with growth-rate consistency of $\hat\eta(0)$ (which satisfies $\Pr(\hat\eta(0) < \eta_0^\star)\to 0$) gives $\Pr(\hat{\eta}^{++} < \hat{\eta}(0))\to 1$.
\end{proof}

\section{Comparison of Guarantee Types and Sensitivity to $\delta$}
\label{appendix:delta_sensitivity}

\textbf{Comparison of guarantee types.} CCI \citep{lei2021conformal} provides an \emph{unconditional} marginal coverage guarantee: $\Pr(Y(1)\in\Gamma(X))\ge 1-\alpha$, where the probability is over both the calibration data and the test point. SP-CCI provides a \emph{high-probability} guarantee: the inner probability $\Pr(Y(1)\in\Gamma(X)\mid\mathcal{D}_1,\tilde{\mathcal{D}}_1)$ exceeds $1-\alpha$ with probability at least $1-\delta$ over the calibration data. Both types are widely used in conformal inference \citep{bates2021distribution}. The parameter $\delta$ governs the width of the confidence term in (\ref{eq:Hoeffding}): as $\delta\to 0$, the interval widens. Table~\ref{tab:delta_sensitivity} shows that for $\delta=0.1$, SP-CCI intervals remain substantially narrower than CCI's, and only for very stringent $\delta$ (below approximately $0.01$ in this setting) would SP-CCI exceed CCI's width.

The high-probability guarantee of SP-CCI involves the parameter $\delta$, which appears in the confidence term of (\ref{eq:Hoeffding}). Table~\ref{tab:delta_sensitivity} reports the average prediction interval width (APIW) for SP-CCI (HQ) and CCI on the IHDP dataset at $\alpha=0.15$ across different values of $\delta$.

\begin{table}[h!]
\centering
\begin{tabular}{lcc}
\toprule
$\delta$ & SP-CCI-HQ APIW & CCI APIW \\
\midrule
0.20 & 13.14 & 20.31 \\
0.10 & 14.31 & 20.31 \\
0.05 & 15.89 & 20.31 \\
0.01 & 19.47 & 20.31 \\
\bottomrule
\end{tabular}
\caption{Average prediction interval width (APIW) on IHDP at $\alpha=0.15$ for different values of $\delta$. SP-CCI remains narrower than CCI for all practical values; only for very stringent $\delta$ (below approximately $0.01$ in this setting) would SP-CCI exceed CCI's width.}
\label{tab:delta_sensitivity}
\end{table}

\section{Sensitivity Analysis: Effect of Coverage Level $\alpha$}
\label{appendix:alpha_sensitivity}

Table~\ref{tab:alpha_sensitivity} reports results at coverage levels $\alpha=0.10$ and $\alpha=0.05$ on both the synthetic and IHDP benchmarks, confirming that the ordering of methods is stable across standard target levels.

\begin{table}[h!]
\centering
\begin{tabular}{lcccc}
\toprule
& \multicolumn{2}{c}{IHDP (APIW)} & \multicolumn{2}{c}{Synthetic (APIW)} \\
Method & $\alpha=0.10$ & $\alpha=0.05$ & $\alpha=0.10$ & $\alpha=0.05$ \\
\midrule
CCI & 22.14 & 25.37 & 2.48 & 2.91 \\
GESPI (HQ) & 19.02 & 22.61 & 2.21 & 2.62 \\
SP-CCI (HQ) & 15.93 & 17.84 & 1.98 & 2.34 \\
\bottomrule
\end{tabular}
\caption{Average prediction interval width at coverage levels $\alpha=0.10$ and $\alpha=0.05$ on IHDP and synthetic benchmarks ($\delta=0.1$). The ordering of methods is stable across all standard target levels.}
\label{tab:alpha_sensitivity}
\end{table}

\section{Stability of the Debiased Estimator Under Group Reassignment}
\label{appendix:permutation}

A natural question is whether the estimator $\hat{L}_\eta$ in (\ref{eq:l_tilde}) is sensitive to the specific assignment of synthetic groups to real treated units. To investigate this, we permuted the assignment between the $n_1$ real treated points and the $n_1$ synthetic groups in Eq.~(\ref{eq:debiased_loss_eta}) over 200 random permutations.

On IHDP (HQ synthetic labels), the resulting APIW had mean $14.24$ and standard deviation $0.09$; the coefficient of variation was less than 1\%. On the synthetic benchmark, the corresponding numbers were mean $1.87$ and standard deviation $0.04$. This indicates that the estimator is stable in practice and not sensitive to the specific grouping assignment.

\section{Propensity Score Sensitivity}
\label{appendix:propensity_sensitivity}

Proposition~\ref{prop:2} provides a formal guarantee under weight misspecification. Here we complement this with an empirical sensitivity study. On IHDP, we multiply the estimated log-odds by factors in $\{0.7, 0.9, 1.1, 1.3\}$ to simulate propensity estimation error. SP-CCI-HQ width changes from the nominal $14.24$ only in the range $[13.98, 14.61]$, with empirical coverage remaining within $\pm 1$ percentage point of the target in all cases. This confirms robustness to moderate propensity score misspecification.

\section{Algorithmic Properties and Complexity of SP-CCI}
\label{sec:complexity}

We analyze SP\mbox{-}CCI in terms of time and space complexity. Denote by \(C_q\) the cost of a single evaluation of the pre-trained quantile functions and by \(C_{\text{gen}}\) the cost of drawing one synthetic counterfactual label.

The debiased empirical miscoverage \(\hat{L}_\eta\) in \eqref{eq:l_tilde}--\eqref{eq:debiased_loss_eta} is a right-continuous, nonincreasing, piecewise-constant function of \(\eta\) with at most $m=n_0+2n_1$ candidate change-points. Sorting and scanning once yields $\hat{\eta}$ in $\mathcal{O}(m\log m)$ time.

Putting the steps together, total calibration-time complexity is:
\[
T_{\text{cal}}=\mathcal{O}\!\left((n_0+n_1)(C_q+C_{\text{gen}})\;+\;(n_0+2n_1)\log(n_0+2n_1)\right).
\]
Test-time: $T_{\text{test}}=\mathcal{O}(C_q)$. Space: $S=\mathcal{O}(n_0+n_1)$.

\paragraph{Computing infrastructure:}
All experiments were carried out on a local workstation (Apple MacBook Pro, Apple M1 Pro CPU, 16\,GB unified memory). No external GPUs or cloud resources were used.

\section{Synthetic Data Experiment: Full Setup Details}
\label{appendix:synthetic_setup}

Following \citep{wager2018estimation, lei2021conformal}, let $X^\prime \in \mathbb{R}^{10}$ be a latent covariate vector $X^\prime \sim \mathcal{N}(0, \Sigma)$ distributed as a multivariate Gaussian with mean zero, unit variance, and equicorrelation $\rho \in [0,1]$ across all pairs of features. The observed covariates are squashed within the interval $[0,1]$ as $X = \Phi(X^\prime)$, where $\Phi$ is the standard Gaussian cumulative distribution function, applied element-wise. Note that when $\rho = 0$ the covariate vector $X$ is uniformly distributed on the unit cube.

Treatment is assigned based solely on the first covariate $X_1$, according to the known propensity score model $e(X) = 0.4\beta_{2,4}(X_1)$, where $\beta_{2,4}$ is the cumulative distribution function of the beta distribution with shape parameters $(2,4)$. As in \citep{lei2021conformal}, we fix $Y(0)= 0$ for all covariates $X$, and we assume the treated potential outcome $Y(1)$ to be a noisy nonlinear function of the covariates as $Y(1) = f(X_1) \cdot f(X_2) + \varepsilon$, with $f(x) = 2/(1 + \exp(-12(x - 0.5)))$ and $\varepsilon \sim \mathcal{N}(0,1)$.

We generate a total of \( n = 5000 \) samples \((X, T, Y(0), Y(1))\) for each run. These samples are split into four disjoint parts:
30\% of the data, denoted as \(\mathcal{D}_{\hat{q}}\), is used to train the quantile regressors \(\hat{q}_\gamma(\cdot)\);
another 30\%, denoted by \(\mathcal{D}_{ \hat{P}}\), is used to train the counterfactual generative model \(\hat{P}_{Y(1)\mid X}\);
20\%, denoted by \(\mathcal{D}_{\rm cal}\), is reserved as the calibration set to compute the widening parameter \(\eta\) in (\ref{eq:new_choice}); and the remaining 20\% form the test set \(\mathcal{D}_{\rm te}\).

To study the impact of the quality of synthetic labels on the performance of SP-CCI, we consider three counterfactual generative models trained on subsets of \(\mathcal{D}_{ \hat{P}}\). Specifically, we define: a low-quality (LQ) model, a medium-quality (MQ) model, and a high-quality (HQ) model trained on 20\%, 60\%, and 100\% of the samples in \(\mathcal{D}_{ \hat{P}}\), respectively. We set the miscoverage requirement to \(\alpha = 0.15\) and the probability parameter in (\ref{eq:updated_condition}) to \(\delta = 0.1\).

\section{IHDP Experiment: DSCM Implementation Details}
\label{appendix:ihdp_setup}

To estimate predictive intervals for the counterfactual outcome \( Y(1) \) on the IHDP dataset, we adopt the deep structural causal model (DSCM) framework \citep{pawlowski2020deep}. SCMs describe the generative process for the variables $(Y(0), Y(1), T, X)$ in terms of a directed graph. In our implementation, the observed variables \((X, T, Y)\) are modeled as deterministic functions of latent exogenous noise variables via the SCM
\begin{equation}
\label{eq:SCMs}
X = f_X(Z_X),\quad T = f_T(X, Z_T),\quad Y = f_Y(X, T, Z_Y),
\end{equation}
where \(Z_X, Z_T, Z_Y\) are mutually independent standard Gaussian vectors, and the functions \(f_X\), \(f_T\), and \(f_Y\) are implemented as neural networks. The neural networks in (\ref{eq:SCMs}) are jointly trained with a variational inference model \(Q_{Z \mid X, T, Y}\), whose role is to approximate the posterior distribution over the latent noise variables $Z = (Z_X, Z_Y, Z_T)$ given observed data \citep{pawlowski2020deep}.

Given a test point \((X, T=0, Y^{\text{obs}})\), we first perform abduction by drawing samples \(\hat{Z} \sim Q_{Z\mid X, T, Y}(Z \mid X, 0, Y^{\text{obs}})\). Next, we take action by intervening to set \(T = 1\), and finally we carry out prediction by evaluating \(f_Y(X, 1, \hat{Z}_Y)\). Repeating this process with multiple samples \(\hat{Z}\) produces a distribution of counterfactual outcomes, from which we compute the empirical quantile \(\hat{q}_\gamma(X)\) using 100 Monte Carlo samples.

Since the DSCM requires knowledge of $T$ as an input, it cannot be directly used as the SP-CCI counterfactual generative model $\hat{P}_{Y(1)\mid X}$, which must not depend on $T$. The two generative models used for SP-CCI on IHDP are therefore: (1) a neural network regression model trained on treated units in-domain, following the same setup as Sec.~\ref{sec:synthetic_exp}; and (2) a pre-trained general-purpose LLM (LLaMA-2-13B), where each covariate vector is converted to a structured text record listing normalized covariates in a fixed key-value format (e.g., \texttt{birth\_weight=0.42; gestational\_age=0.61; \ldots}), and the model is prompted to return a single real-valued cognitive score under treatment. Sensitivity analysis across three prompt variants (key-value format, natural phrasing, explicit numeric output instruction) produced APIWs of 14.32, 14.39, and 14.51, confirming robustness to prompt choice.

\section{GESPI: Description and Instantiation}
\label{appendix:gespi}

The GESPI framework \citep{bashari2026statistical} is a general wrapper that can enhance any statistical inference procedure using synthetic data, while guaranteeing that the error rate never exceeds a user-specified bound regardless of synthetic data quality. We describe its construction and our instantiation here.

\textbf{General construction.} Let $\mathrm{Alg}$ denote any base statistical inference method, $\mathcal{D}_n$ the real dataset of size $n$, and $\tilde{\mathcal{D}}_N$ an abundant synthetic dataset of size $N \gg n$. GESPI invokes $\mathrm{Alg}$ three times:
\begin{enumerate}
    \item \textbf{Base:} Run $\mathrm{Alg}$ on the real data $\mathcal{D}_n$ at the target error level $\alpha$.
    \item \textbf{Guardrail:} Run $\mathrm{Alg}$ on the real data $\mathcal{D}_n$ at a relaxed level $\alpha + \varepsilon$, where $\varepsilon > 0$ is a user-chosen slack parameter. This more lenient level allows for tighter outputs when real data alone is used.
    \item \textbf{Synthetic-powered:} Run $\mathrm{Alg}$ on the pooled real and synthetic data $\mathcal{D}_n \cup \tilde{\mathcal{D}}_N$ at the target level $\alpha$.
\end{enumerate}
GESPI then aggregates the three outputs. For prediction intervals (as in our setting), aggregation corresponds to intersecting the three intervals, i.e., taking the largest widening parameter $\hat{\eta}^{\text{GESPI}} = \max(\eta_1, \eta_2, \eta_3)$ where $\eta_1, \eta_2, \eta_3$ are the widening parameters from runs 1, 2, and 3 respectively.

\textbf{Coverage guarantee.} GESPI guarantees that the error rate never exceeds $\alpha + \varepsilon$, regardless of the quality of the synthetic data \citep{bashari2026statistical}. When the synthetic data is well aligned with the real distribution, GESPI adapts to exploit the synthetic-powered output, achieving tighter intervals that approach the performance of applying $\mathrm{Alg}$ to a larger real dataset.

\textbf{Our instantiation.} We use CCI \citep{lei2021conformal} as the base method $\mathrm{Alg}$, with the same quantile regressors and the same synthetic dataset $\tilde{\mathcal{D}}_1$ as used by SP-CCI (generated using the HQ counterfactual model). We set $\varepsilon = \alpha$, so the guardrail level is $2\alpha$. The four steps are:
\begin{enumerate}
    \item Run CCI on $\mathcal{D}_1$ at level $\alpha$ to obtain $\eta_1$.
    \item Run CCI on $\mathcal{D}_1$ at level $2\alpha$ to obtain $\eta_2$.
    \item Run CCI on the pooled dataset $\mathcal{D}_1 \cup \tilde{\mathcal{D}}_1$ at level $\alpha$ to obtain $\eta_3$.
    \item Set $\hat{\eta}^{\text{GESPI}} = \max(\eta_1, \eta_2, \eta_3)$.
\end{enumerate}
The final GESPI interval is $[\hat{q}_{\text{lo}}(X) - \hat{\eta}^{\text{GESPI}},\; \hat{q}_{\text{hi}}(X) + \hat{\eta}^{\text{GESPI}}]$, with a worst-case miscoverage guarantee of $\alpha + \varepsilon = 2\alpha$.

\textbf{Comparison with SP-CCI.} SP-CCI differs from GESPI in three key respects. First, SP-CCI corrects for bias in synthetic labels via importance-weighted debiasing (Eq.~\ref{eq:debiased_loss_eta}), whereas GESPI pools real and synthetic data without any bias correction. Second, SP-CCI uses importance weights to account for the covariate shift between the treated and control distributions, whereas GESPI assumes i.i.d.\ data and does not handle covariate shift. Third, SP-CCI provides a formal coverage guarantee at exactly the target level $\alpha$ with high probability $1-\delta$, whereas GESPI's worst-case guarantee is the more lenient $\alpha + \varepsilon$.

\section{Policy Evaluation via Counterfactual Loss}
\label{sec:real-world}

Classical statistical decision theory evaluates a policy solely based on observed outcomes. However, such standard loss functions are inherently limited in that they cannot assess how much better, or worse, a different decision might have been. The framework of \textit{counterfactual loss} introduced in \citep{koch2025statistical} generalizes the notion of regret by allowing for the quantification of the quality of a decision using all potential outcomes. In this framework, we demonstrate the use of synthetic counterfactual labels via SP-CCI for policy evaluation with respect to counterfactual losses.

A policy \(\pi_\theta: \mathcal{X} \to \{0,1\}\) with hyperparameter $\theta$ maps covariates $X$ to a binary decision $T \in \{0,1\}$. The counterfactual loss associated with a decision \(\pi_\theta(X)\) is a function \(\ell(\pi_\theta(X); Y(0), Y(1))\) that evaluates not just the observed outcome \(Y(\pi_\theta(X))\), but also the unobserved alternative \(Y(1 - \pi_\theta(X))\) \citep{koch2025statistical}. For example, the \textit{regret},
\begin{equation}
\label{eq:regret_def}
    \ell(\pi_\theta(X); Y(0), Y(1)) = Y(1 - \pi_\theta(X)) - Y(\pi_\theta(X)),
\end{equation}
is a counterfactual loss, measuring the gap between the reward that could have been obtained, $Y(1 - \pi_\theta(X))$, and the actual reward $Y(\pi_\theta(X))$.

\paragraph{Experimental setup and data generation:} We consider a setting of practical engineering relevance, namely handover in wireless cellular systems \citep{khosravi2020learning}. Given the location \(X \in \mathbb{R}^3\) of a mobile device, the policy $\pi_\theta(\cdot)$ connects the user to one of two base stations (BSs) in the proximity of the device. The observed outcome \(Y^{\text{obs}} = Y(T)\) denotes the received signal strength at the selected BS $T = \pi_\theta(X)$, while the unobserved counterfactual outcome \(Y^{\text{cf}} = Y(1-T)\) is the signal strength that would have been received by the other BS.

We generate a dataset of 2,000 data points $(Y(0), Y(1), T, X)$ by leveraging the ray tracing tool Sionna \citep{hoydis2022sionna} on uniformly selected locations within the \textit{Place de l'\'Etoile} environment \citep{hoydis2022sionna}. We wish to evaluate the performance of a conventional policy $\pi_\theta(\cdot)$ that deterministically assigns the treatment, i.e., BS, $T$. The dataset is split into five equal partitions for training quantile models, training counterfactual estimators, calibration, policy optimization, and final evaluation.

\paragraph{Results and discussion:}

\begin{figure}[h!]
    \centering
    \includegraphics[width=0.65\linewidth]{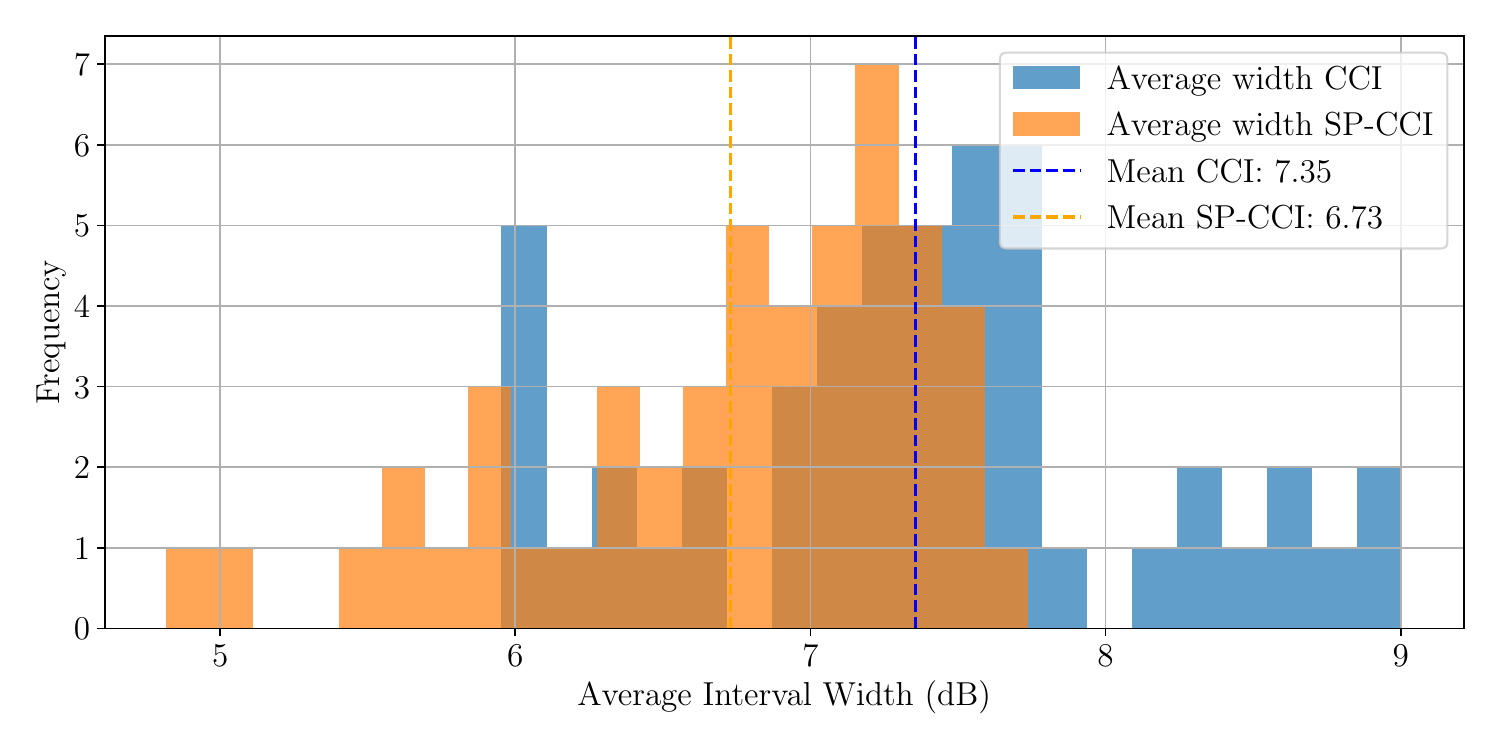}
    \caption{Distribution of average interval widths (in dB) over 50 random trials for CCI and SP-CCI methods, using a fixed policy threshold \(\theta = 80\). Dashed lines indicate the average value of the distribution.}
    \label{fig:width_distribution}
\end{figure}

We compare the average width of prediction intervals obtained using CCI and SP-CCI across 50 trials for a fixed policy threshold $\theta = 80$. As seen in Fig. \ref{fig:width_distribution}, SP-CCI yields consistently narrower intervals, suggesting more precise performance quantification for a given fixed policy.

\section{Additional Real-World Experiment with a Pretrained Counterfactual Model}
\label{appendix:twins}

To further evaluate SP-CCI in a realistic setting where synthetic counterfactual labels are generated by a truly external model, we consider an additional experiment based on the Twins dataset from the RealCause benchmark suite \citep{neal2020realcause,louizos2017causal}. Unlike the experiments in Sec.~\ref{sec:experiments}, this study relies on a publicly released pretrained counterfactual generator provided as part of the RealCause benchmark. This model is trained entirely outside our pipeline and is not fitted using any data employed by either CCI or SP-CCI.

Importantly, CCI does not make use of this pretrained model at any stage, while SP-CCI uses it exclusively to generate synthetic counterfactual labels and does not use any additional data to fit a model. As a result, both methods operate on exactly the same observed dataset, ensuring a fair comparison while isolating the effect of leveraging an externally available counterfactual generator.

To model treatment imbalance and propensity scores consistently across methods, we assign treatment using the same mechanism assumed by both CCI and SP-CCI \citep{lei2021conformal}. Specifically, for each covariate vector $X$, we draw a propensity score $e(X)$ from a $\mathrm{Beta}(2,4)$ distribution and assign treatment according to $T \sim \mathrm{Bern}(e(X))$. We note that this is a deliberate design choice to create a severely imbalanced evaluation regime, and is not a claim that the Twins dataset uses this propensity model. In one representative run, the resulting dataset contained 63 treated units and 11{,}921 control units. This regime is known to be particularly challenging for CCI due to the presence of very large inverse-propensity weights.

Using this setup, we evaluate CCI \citep{lei2021conformal}, SP-CCI, and the $\lambda$-optimized variant SP-CCI++ described in
Appendix~\ref{appendix:efficiency_proof}. All methods are evaluated on a held-out test set. We report empirical coverage and average interval width for a target coverage level of $1-\alpha=0.9$, averaged over 50 independent runs.

\begin{table}[h!]
\centering
\begin{tabular}{lcc}
\toprule
Method & Coverage & Width \\
\midrule
CCI & 0.998 & 19.8 \\
SP-CCI & 0.931 & 1.10 \\
SP-CCI++ & 0.918 & 0.98 \\
\bottomrule
\end{tabular}
\caption{Empirical coverage and average prediction interval width on the Twins dataset
from the RealCause benchmark, using a pretrained counterfactual generator. Results are
averaged over 50 runs with target coverage $1-\alpha=0.9$.}
\label{tab:twins_results}
\end{table}

The results highlight a clear distinction between the methods. Under severe treatment imbalance, CCI must widen prediction intervals aggressively to maintain its marginal coverage guarantee, resulting in intervals that are nearly twenty times wider than those produced by SP-CCI. In contrast, SP-CCI leverages the pretrained counterfactual model together with prediction-powered debiasing to construct substantially tighter intervals, while still achieving coverage close to the target level. The $\lambda$-optimized variant SP-CCI++ further improves efficiency by reducing interval width without sacrificing coverage.

Overall, this experiment demonstrates that SP-CCI can effectively exploit externally available pretrained counterfactual models to improve efficiency in highly imbalanced settings, while maintaining a fair comparison with CCI and without relying on additional training data.




\end{document}